\documentclass[sigconf]{acmart}

\AtBeginDocument{%
  \providecommand\BibTeX{{%
    \normalfont B\kern-0.5em{\scshape i\kern-0.25em b}\kern-0.8em\TeX}}}

 \renewcommand\footnotetextcopyrightpermission[1]{} 
 \setcopyright{none}

\usepackage{times}
\usepackage{amsmath}
\usepackage{latexsym}
\usepackage{graphicx}
\usepackage{tabularx}
\usepackage{makecell}
\usepackage{color}
\usepackage{algorithm}
\usepackage{algorithmic}
\usepackage{multicol}
\usepackage{multirow}
\usepackage{array}
\usepackage{adjustbox}
\usepackage{tabularx}
\usepackage{booktabs}

\begin{document}

\title{Quantity Matters: Towards Assessing and Mitigating Number Hallucination in Large Vision-Language Models}


%
%


\author{Huixuan Zhang}
\affiliation{%
  \institution{Wangxuan Institute of Computer Technology, Peking University}
  \city{Beijing}
  \country{China}}
\email{zhanghuixuan@stu.pku.edu.cn}

\author{Junzhe Zhang}
\affiliation{%
  \institution{Wangxuan Institute of Computer Technology, Peking University}
  \city{Beijing}
  \country{China}}
\email{junzhezhang@stu.pku.edu.cn}

\author{Xiaojun Wan}
\affiliation{%
  \institution{Wangxuan Institute of Computer Technology, Peking University}
  \city{Beijing}
  \country{China}}
\email{wanxiaojun@pku.edu.cn}
\begin{abstract}
Large-scale vision-language models have demonstrated impressive skill in handling tasks that involve both areas. Nevertheless, these models frequently experience significant issues with generating inaccurate information, which is hallucination. In this study, we concentrate on a specific type of hallucination—number hallucination, referring to models incorrectly identifying the number of certain objects in pictures. We perform quantitative evaluations regarding number hallucination, showing it to be critical in major open-source large vision-language models. Furthermore, we utilizes two related tasks to conduct an in-depth analysis of number hallucination, revealing the severe inner and outer inconsistency among all tasks. Based on this examination, we devise a training approach aimed at improving consistency to reduce number hallucinations, which leads to an 8\% enhancement in performance over direct finetuning methods. Our code and dataset will be released to the community.
\end{abstract}



\maketitle

\section{Introduction}
Large language models (LLMs) have demonstrated remarkable capabilities in solving diverse problems by adhering to human instructions. Consequently, researchers have undertaken numerous endeavors to integrate multiple modalities into a single large language model, with the most successful being large vision-language models (LVLMs). LVLMs are designed to process both images and text as input and generate text-only responses. Among all LVLMs, GPT4-V\citep{openai2023gpt4}stands out as the most powerful one, showcasing extraordinary proficiency in addressing various vision-text challenges, including tasks such as vision question answering (VQA) and image captioning.

Nevertheless, despite the success of large vision-language models, they still grapple with several challenges, with hallucination being the most severe among them. In this context, hallucination primarily pertains to the discrepancy between the model's output and the content present in the input image. Among various forms of hallucination, object hallucination, which centers on determining whether an object is present in the image, has been extensively discussed. Numerous methods for evaluation and mitigation have been proposed to tackle this particular challenge(\citet{zhou2023analyzingobject}, \citet{leng2023mitigatingobject}, \citet{zhai2023halle}). The majority of prior studies employ binary classification questions (answered with `yes' or `no') to assess the model's performance(\citet{li2023evaluating}, \citet{wang2023mitigating}).

While simple object hallucination has been extensively studied, a more nuanced and broadly applicable discussion on object hallucination is still lacking. For a comprehensive understanding of an image, the model must not only discern the presence of a specific object but also accurately determine the quantity of that particular object. Without this capability, the model cannot fully comprehend the image, thereby leaving the door open to potential risks of other forms of hallucination.

In this study, we focus on investigating a new form of object hallucination, specifically termed as number hallucination. Number hallucination occurs when models are unable to accurately identify the correct quantity of a particular type of object in an image. Table \ref{tab:1} illustrates two typical examples of number hallucination. We also conduct detailed experiments and point that using binary classification questions is not enough to fully assess the model's performance. Instead, we choose to use a direct inquiry to the model about the number of a specific object in an image, utilizing prompts to control its output. This approach is more aligned with real-world scenarios and facilitates a more quantitative evaluation of model performance.

\begin{table}[htbp]
    \centering
    \begin{tabular}{c|c}
        \toprule[1.5pt]
      Input Picture  &  Prompt and Output \\
      \midrule[1pt]
      \makecell[c]{
      \includegraphics[width = 3.0cm, height = 3.5cm]{latex/fig1.jpg}
      }
         & \makecell{Prompt: How many  \textcolor{green}{motorcycles} \\are  there in this picture? \\ Answer in a  single number. \\
         Output: \textcolor{red}{2} \\
         Gold Answer: \textcolor{green}{1}} \\
       \hline
       \makecell[c]{
       \includegraphics[width = 3.0cm, height = 3.5cm]{latex/fig1b.jpg}
       } & \makecell{Prompt: How many  \textcolor{green}{persons} \\are there in this picture? \\ Answer in a  single number. \\
         Output: \textcolor{red}{3} \\
         Gold Answer: \textcolor{green}{4}}\\
       \bottomrule[1.5pt]
    \end{tabular}
    \caption{Two examples of number hallucination.}
    \label{tab:1}
\end{table}

Furthermore, in the context of the counting task, there exist multiple related tasks that help reflect overall performance such as binary classification questions (Section \ref{sec:4.1}) and comparative questions (Section \ref{sec:4.2}). Naturally we would like to investigate the consistency between these tasks to attain a more comprehensive understanding of number hallucination. We find that inconsistency is a common problem that appears despite task format. Building upon these analyses, we introduce a consistency training method to enhance the model's capability. Our findings demonstrate that finetuning the model from different perspectives yields superior results compared to direct finetuning on the single primary task. This methodological insight is intriguing and holds potential for extension to various tasks that necessitate finetuning of LVLMs.

To summarize, our contributions are listed as follows \footnote{Our code and dataset will be released to the community.}:

\begin{itemize}
    \item We introduce a new form of object hallucination, specifically referred to as number hallucination, and present a dataset with 20k pieces of data along with corresponding evaluation metrics and results. We point out that all LVLMs we investigated have an average MAE of around $2$ on this dataset, which is disasterous. To the best of our knowledge, our study represents the first comprehensive and quantitative exploration of this particular type of hallucination.
    \item We conduct a detailed analysis of number hallucinations by exploring the inner and outer inconsistency from different perspectives. We point out that all models suffer from both inner and outer inconsistencies despite task or prompt format, which reveals the model's uncertainty and confusion and is a potential cause of number hallucination. 
    \item We propose a consistency training method which is extendable to any large vision-language model (LVLM). We substantiate the effectiveness of this method through our experiments by averagely 8\% compared with directly finetuning methods.
\end{itemize}

\section{Related Works}
\subsection{Large Vision-Language Models}
With the success of large language models (LLMs), there has been a surge in efforts to incorporate additional modalities into LLMs, leveraging their powerful recognition capabilities. BLIP-2 \citep{li2023blip}, for instance, introduces the Q-former to align image input into tokens suitable for utilization by LLMs. LLaVA \citep{liu2023llava} and LLaVA-v1.5 \citep{liu2023improvedllava} employ visual instruction tuning and leverage simple linear layers or multi-layer perceptrons to map image output into the text feature space. InstructBLIP \citep{instructblip} achieves instruction following abilities by performing instruction tuning on the BLIP-2 backbone. Notably, MiniGPT-4 \citep{zhu2023minigpt} and MiniGPT-v2 \citep{chen2023minigptv2} are also powerful LVLMs that exhibit strong performance across various vision-language tasks. Among all open-source and closed-source LVLMs, GPT-4V \citep{openai2023gpt4} retains its status as the most powerful one.

Hallucination is an extensively discussed phenomenon and numerous studies have delved into the examination of hallucinations in large vision-language models (LVLMs). The most extensively discussed form of hallucination is object hallucination, centering around the determination of whether a specific object exists in a given image. In the context of caption-based evaluation, CHAIR \citep{rohrbach2018chair} stands out as a classic metric that compares objects mentioned in the caption with those present in the provided image. POPE \citep{li2023evaluating}, on the other hand, employs a binary classification approach by directly asking the model whether a particular object exists in the provided image. \citet{zhou2023analyzingobject} conducted an analysis of factors influencing object hallucination and introduced LURE. \citet{leng2023mitigatingobject} proposed visual contrastive decoding to mitigate object hallucinations, while \citet{zhai2023halle} suggested a switch-based method to address this issue. However, these studies predominantly focus on the mere existence of objects, which may not be sufficient for gauging whether the model truly comprehends the image.

Beyond object hallucination, there are also studies addressing more generalized forms of hallucinations. \citet{gunjal2023detecting} proposed a dataset and reward model to assess and mitigate hallucinations in captions. However, many caption-based hallucination evaluation methods require a lot of training modules, so their performance and stability remain uncertain. RAH \citep{chen2023mitigating} and FGHE \citep{wang2023mitigating} are two datasets similar to POPE that can be employed for evaluating fine-grained hallucinations.  \citet{wang2023evaluation} and \citet{jing2023faithscore} utilizes LLMs and LVLMs to assist detection of hallucination, but LLMs and LVLMs bear hallucinations themselves, making the evaluation results less reliable. POPE-like evaluation metrics rely on binary classification questions and might not fully capture the model's ability, making them insufficient for quantitative comparisons.

\subsection{Counting Task in VQA}

There have been studies specifically addressing the task of counting objects in visual question-answering. Earlier works such as \citet{nips2010count} and \citet{Segui2015count} laid some groundwork in this area. \citet{zhang2018learning} proposed a counting module. Additionally, \citet{Liu2020count} introduced a blockwise counting method to tackle issues like inaccurate generated regression targets and imbalanced samples, which were significant challenges in previous counting methods. However, these methods often require specific designs for the task and modifications to the entire model architecture, which may be insufficient and inefficient in the context of large models nowadays.

For datasets and benchmarks, HowmanyQA \citep{trott2017interpretable} and TallyQA \citep{acharya2019tallyqa} are two datasets focusing on counting tasks. However, these proposed datasets both bear the problem that both ground truth labels and the queried objects are not uniformly distributed, resulting in probably biased evaluation results. What's more, these works choose accuracy as one evaluation metric, which is not suitable for non-uniformly distributed labels. Besides, these datasets are directly constructed in the form of VQA, making it hard to be transformed into other complex formats like what we have done in Section \ref{sec:4.2} (because there may be only one counting question regarding one image under most circumstances). 

\section{Evaluating Number Hallucination in LVLMs}
In this section, we begin by providing a concise definition of number hallucination in large vision-language models (LVLMs) and outline our evaluation settings for assessing this phenomenon. Subsequently, we present the results of our evaluation and provide brief discussions on the findings.

\subsection{Task Definition}
\label{sec:3.1}
Number hallucination occurs when models fail to accurately determine the correct quantity of a specific object category in a given image.  Formally, for an image $i$ and given object $c$, let $n_{c}$ denote the ground truth number of $c$ visible in $i$. If a model responds with $n^{'}( \neq n_{c})$ $c$ in $i$ (in any format), we classify the model as suffering from number hallucination.

As observed, number hallucination can manifest in various vision-language tasks, including image captioning and vision-question answering (VQA). For the sake of simplicity in quantitative evaluation, we opt for VQA as the foundational format for assessing number hallucinations. For any given image $i$ and object $c$, the question is formatted as:

\textit{Prompt: How many $c$ are there in this picture? Answer in a single number.}

This task format offers several advantages. In comparison to caption-based evaluation methods, this format makes it considerably easier to extract meaningful responses (numbers) and provides a clearer indication of the model's performance. Captions may segregate the same object into different sentences or overlook existing objects, even when the model has detected them. These complexities in captions add extra challenges to the evaluation of number hallucinations.

In contrast to binary classification-based methods, firstly, this approach aligns more closely with real-life scenarios and provides a clearer insight into the model's view on this task. Secondly, as can be seen in Table \ref{tab:3}, all models exhibit a notable inclination to answer `Yes' more frequently than `No'. This observation underscores the inherent bias of all models toward answering `Yes', making evaluation methods employing binary classification questions less reliable. Binary classification questions also suffer from severe inconsistency problems, which will be further discussed in Section \ref{sec:4.1}. Lastly, binary classification questions fail to capture the severity of hallucinations. For instance, mistaking $4$ cats as $3$ cats may be relatively acceptable, but mistaking $4$ cats as $10$ cats is disastrous and entirely unacceptable. It is difficult for binary classification questions to delineate this fine-grained difference.

\subsection{Dataset and Evaluation Metrics}
\paragraph{Dataset}
\label{sec:3.2}
MSCOCO \citep{lin2014mscoco} is a large-scale object detection, segmentation, and captioning dataset. All tests in this study are conducted on the MSCOCO2014 validation set.\footnote{Test set is not used because we cannot find annotation results on the official website.} Leveraging the provided detection and segmentation labels, we can construct a counting VQA dataset. Symbolically, for any given image $i$ with labeled bounding boxes $\{(b_{1},c_{1}),...,(b_{m},c_{m})\}$, where $b_{i}$ represents a bounding box and $c_{i}$ represents its corresponding object, the total number $n_{o}$ of a certain object $o$ in image $i$ can be denoted as:

\[
    n_{o} = \sum_{k = 1}^{m}I[c_{k}=o]
\]

where 

\[
    I[x] = \left\{ \begin{aligned}&1,\ x\ is\ True \\& 0,\ ot herwise \end{aligned} \right.
\]

Following this approach, we can construct a dataset with ground truth where each data instance can be denoted as $(i, c, n_{c})$, with $i$ denoting the given image and $c$ denoting the specified object. However, based on our observation, the distribution of answers is not uniform. For instance, in the original dataset, there are about 68\% questions with answer $n_{c} = 1$. What's more, the distribution of objects is also non-uniform. To address these biases, we propose double k-uniform sampling. The core of this algorithm is limiting the number of data with object $c$ and answer $n_{c} = n$ not exceeding $k$. We present double k-uniform sampling in algorithm \ref{alg:1}. Adjusting the value of $k$ allows us to strike a balance between achieving a more uniform distribution and retaining a larger dataset. In the subsequent discussion, we opt for $k = 50$. 

\begin{algorithm}
    \caption{Double k-Uniform Sampling}             \begin{algorithmic}[1]
        \STATE Input dataset $S = \{(i, c, n_{c})\}$ and $k$.
        \STATE Randomly shuffle the dataset $S$.
        \STATE Initialize $m(c,n) \leftarrow 0, S^{'} = \phi$ 
        \FOR{$j = 1, 2..., |S|$}
            \IF{$m(c_{j}, n_{cj}) > k$}
                \STATE continue
            \ENDIF
            \STATE $S^{'} \leftarrow S^{'}\cup \{(i_{j},c_{j},n_{cj})\}$
            \STATE $m(c_{j},n_{cj})\leftarrow m(c_{j},n_{cj}) +1$
        \ENDFOR
        \STATE Output $S^{'}$
    \end{algorithmic}
    \label{alg:1}
\end{algorithm} 

It is worth mentioning that, when constructing dataset, we only retain data $(i,c,n_{c})$ with $n_{c} > 0$. There are two main considerations for this choice. First of all, there are many objects that do not appear in a given image and retaining all these questions is too expensive. If we perform k-uniform sampling similarly, this may result in random selection, which is insufficient for reflecting the model's performance according to \citep{li2023evaluating}. More importantly, we conduct an experiment on LLaVA with randomly selected $(i,c,n_{c})$ with $n_{c} = 0$ and find out that LLaVA correctly answers 95\% of these questions, which meets the its high POPE evaluation result \citep{liu2023improvedllava}. Based on these considerations, we only retain data $(i,c,n_{c})$ with $n_{c} > 0$ to focus on the number of an existing object.

Another consideration is that, one bounding box in MSCOCO annotation may not exactly correspond to one existing object. Luckily, we found one solution of utilizing the `is\_crowd' field of the annotation. When constructing original dataset $S$, if there exists one annotation $(b_{j}, c_{j})$ in image $i$ with `is\_crowd' being True, the data point $(i, c_{j}, n_{c_{j}})$ is not included in original dataset $S$. This approach ensures the calculation of objects in the original dataset $S$ is correct.

\begin{figure}[htbp]
    \centering
    \includegraphics[width=0.45\textwidth]{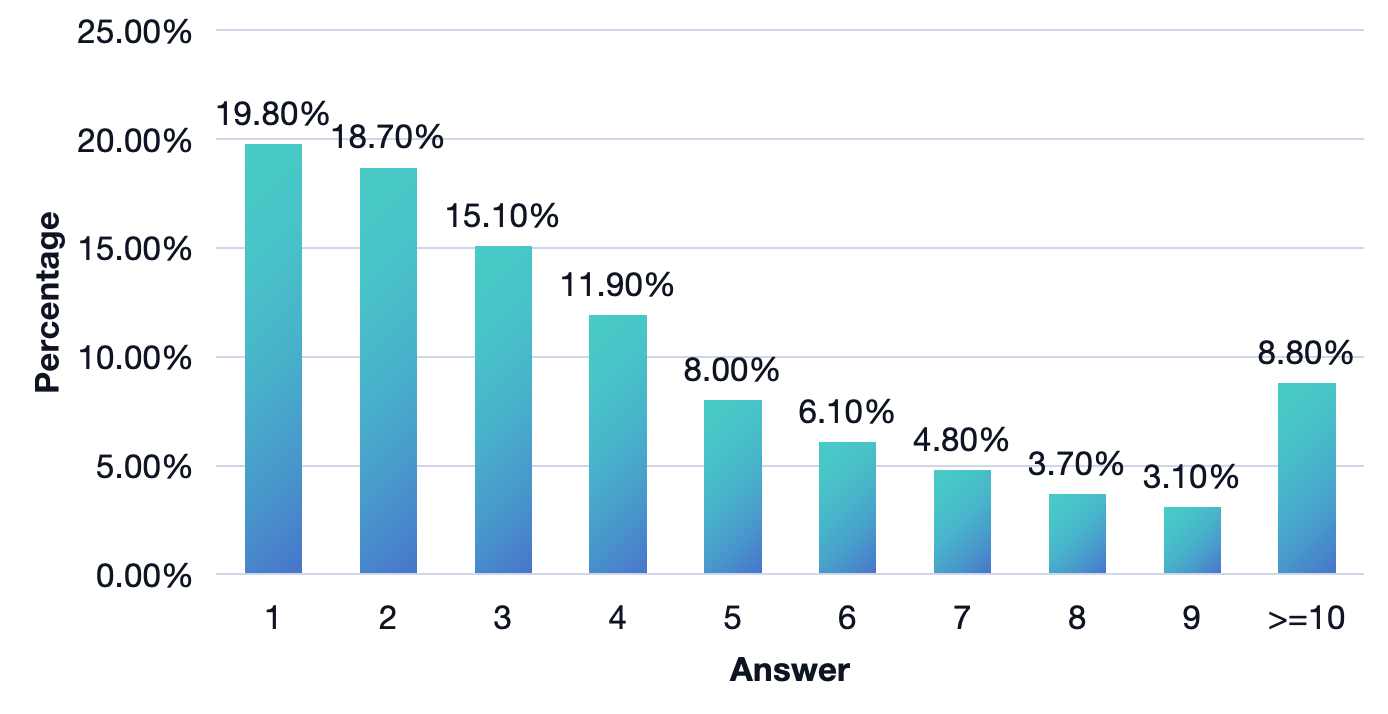}    
    \caption{The distribution of the answers to the questions.}
    \label{tab:app-1}
\end{figure}

After performing $k$-uniform sampling, we manually sampled and checked some labels in the final $S^{'}$ to ensure the quality of the final dataset. Finally we obtain a dataset with 20200 pieces of data. The distribution of labels are listed in Figure \ref{tab:app-1}.

As can be seen from the Figure \ref{tab:app-1}, though k-uniform sampling is performed, questions with answer $\geq 10$ is still too less. Therefore to avoid randomness, when calculating F-scores, questions with answers $\geq 10$ are considered one category.

\paragraph{Evaluation Metrics and Decoding Strategy}
In evaluating the counting task, our focus extends to two perspectives. Firstly, we aim for the model to provide correct answers to as many questions as possible. Secondly, we seek minimal errors. 

To address the first perspective, we treat this problem as a multi-classification task and employ F-scores(macro-F1 and weighted-F1) to evaluate the model's performance. Macro-F1 is more suitable for non-uniformly distributed labels like the ones in our test set because it gives equal weight to both large and small categories. However, since there is less data in some categories, the value of macro-F1 may be more easily affected by the results of these data, which introduces more randomness. To address this problem, weighted-F1 is introduced to offer a more comprehensive understanding of the performance of models. Weighted-F1 gives larger weights to larger categories, which reduces randomness. However, if the model perform much better on larger categories than on smaller ones, weighted-F1 may be cheated. Thus only by combining macro-F1 and weighted-F1 can we fully reveal the capability of models. 

For the second perspective, we use mean absolute error (MAE) as an evaluation metric. The employment of both F-scores and MAE provides a more comprehensive and quantitative understanding of the severity of number hallucination. For instance, a model that never accurately counts objects, even with a small MAE, is considered ineffective. Similarly, a model that occasionally counts objects correctly but often makes significant errors should also be deemed problematic. Only when a model that exhibits both high F-scores and low MAE can it be considered free from number hallucinations.

 We use greedy decoding to make sure no randomness is introduced to the responses. All following evaluations use the same setting. Since the answers to our questions are relatively short, we believe this setting does not harm performance.

\subsection{Evaluation Results}
\label{sec:3.3}
We assess number hallucinations using four of the most advanced large vision-language models (LVLMs), which are LLaVA-v1.5-7B(namely LLaVA), LLaVA-v1.5-13B (namely LLaVA-L)\citep{liu2023improvedllava}, InstructBLIP (\citet{instructblip}, equipped with Vicuna-7B \citep{vicuna2023}) and MiniGPT-v2 (\citet{chen2023minigptv2}, equipped with LLaMA-2-chat-7B \citep{touvron2023llama2}). The results are shown in Table \ref{tab:2}. We also add a random result (randomly select an answer ranging from from $1$ to $10$) as a comparison. Since testing GPT-4V is very expensive, we perform a quantitative evaluation on a small subset of our test set. Details can be found in Appendix A.

\begin{table}[htbp]
    \centering
    \begin{tabular}{c|ccc}
    \hline
        Model & Macro-F1 & Weighted-F1 & MAE \\
        \hline
        Random & 0.086 & 0.103 & 3.56 \\
        LLaVA &  0.186 & 0.288 & 1.84 \\
        LLaVA-L & 0.196 & 0.299 & 1.80 \\
        InstructBLIP & 0.201 & 0.300 & 2.35 \\
        MiniGPT-v2 & 0.193 & 0.293 & 2.06 \\
        \hline 
    \end{tabular}
    \caption{Number hallucination evaluation results of different LVLMs.}
    \label{tab:2}
\end{table}

We can see all models exhibit severe number hallucinations, as their performance over Macro-F1, Weighted-F1 and MAE is generally poor. For instance, they exhibit an average MAE of about $2$, while random answers just bear an MAE of $3.56$. Even LLaVA-v1.5 models, which demonstrate higher POPE \citep{liu2023improvedllava}(indicating less severe object hallucination), still experience number hallucinations similar to other models. This outcome underscores the limitations of traditional object hallucination studies and evaluations, and emphasizes the significance of addressing this number hallucination.

Furthermore, the comparison between LLaVA and LLaVA-L yields a surprising result that larger models do not significantly outperform smaller models. This finding suggests that merely scaling up models might not be an effective strategy for mitigating number hallucination, indicating that the issue may not solely stem from a lack of expressive ability. This observation underscores the need for more detailed and comprehensive studies on number hallucination.

\section{Consistency Analysis from Multiple Perspectives}
\label{sec:4}

In addition to the primary counting task discussed in Section \ref{sec:3.1} (namely primal task), there are also relevant tasks that provide insights into 
number hallucination exhibited by a model. Further, by analyzing the consistency among the outputs of the related tasks and the primal task, we can achieve a more comprehensive understanding of number hallucination. In this context, we mainly analyze two types of consistency, which are inner consistency and outer consistency. Inner consistency represents the consistency of output within the same task format with different settings, while outer consistency represents the consistency between outputs of the primal task and the related task. We take two related tasks and reveal the severe inner and outer inconsistency that both tasks exhibit. This inconsistency is a common problem
shared by both tasks, and is a potential cause of number hallucination.

\subsection{Binary Classification Questions}
\label{sec:4.1}
Binary classification questions are widely used for evaluating hallucinations in LVLMs. In this task, for any given object $c$ and number $n$, the binary classification question can be formatted as:

\textit{Prompt: There is(are) $n$ $c$ in this picture, is that correct? Answer yes or no.}

By modifying $n$, we can generate different questions and explore the inner inconsistency between different settings of $n$. To analyze outer inconsistency, we can specify $n$ as the response to direct asking question (the prompt given in Section \ref{sec:3.1}) of the model. We generate questions using the following three ways:

\begin{itemize}
    \item I: Let $n$ equal to the ground truth.
    \item II: Let $n$ equal to the model's own generated answer.
    \item III: With a 50\% probability, uniformly select $n$ around the ground truth. Otherwise, let $n$ equal to the ground truth. \label{sec:4.1.III}
\end{itemize}

\begin{table}[htbp]
    \centering
    \begin{tabular}{c|ccc}
    \hline
        Model & Prompt & Accuracy & `Yes' Ratio \\
        \hline
        \multirow{3}{*}{LLaVA} & I & 0.749 & 0.749 \\
        ~ & II & 0.437 & 0.843 \\
        ~ & III & 0.649 & 0.604 \\
        \hline
        \multirow{3}{*}{LLaVA-L} & I & 0.755 & 0.755 \\
        ~ & II & 0.365 & 0.920 \\
        ~ & III & 0.604 & 0.656 \\
        \hline
        \multirow{3}{*}{InstructBLIP} & I & 0.992 & 0.992 \\
        ~ & II & 0.346 & 0.948 \\
        ~ & III & 0.513 & 0.984 \\
        \hline
        \multirow{3}{*}{MiniGPT-v2} & I & 0.751 & 0.751 \\
        ~ & II & 0.521 & 0.742 \\
        ~ & III & 0.618 & 0.639 \\
        \hline
    \end{tabular}
    \caption{Evaluation results on binary classification questions. Accuracy is calculated with the ground truth \textit{yes} or \textit{no} answers. `yes' ratio is the ratio of \textit{yes} answers generated by the model.}
    \label{tab:3}
\end{table}

\paragraph{Inner Inconsistency}
In this context, inner inconsistency refers to model responding `yes' to even contradictory questions. An example is presented in Table \ref{tab:4}. Despite the evident contradiction between the two questions (different $n$ setting), the model answer `yes' to both. In fact, by comparing prompt I and prompt II, model shows this inner inconsistency at a ratio of 41\%, which is very high because a model that randomly answers `yes' or `no' will bear an inner inconsistency of just about 25\%. This high ratio reveals a significant challenge that it becomes uncertain what the model's actual understanding is towards the question. Specifically, if the model answers `yes' to a ground truth, we cannot confidently assert that the model precisely knows the number of objects in the given image. Moreover, even if the model answers `no' to a hallucinated expression, it remains unclear whether the model accurately comprehends the ground truth or not. This inner inconsistency suggests that the model struggles to count the number of objects accurately and exhibits confusion, thereby contributing to number hallucination.

\begin{table}[htbp]
    \centering
    \begin{tabular}{c|c}
    \toprule[1.5pt]
    Input Picture & Prompt and Output \\
    \midrule[1pt]
        \multirow{2}{*}{\includegraphics[width = 3.5cm, height = 2.5cm]{latex/fig1.jpg}} & 
        \makecell[c]{Prompt: There are  \textcolor{red}{2} motorcycles \\in this picture, is that correct? \\Answer yes or no. \\ Answer: \textcolor{red}{Yes}} \\ \cline{2-2} ~ & \makecell[c]{Prompt: There is  \textcolor{green}{1} motorcycle \\in this picture, is that correct? \\ Answer yes or no.} \\  & Answer: \textcolor{green}{Yes}\\ 
        \bottomrule[1.5pt]
    \end{tabular}
    \caption{Example of model answering 'yes' to contradictory questions. This example is drawn from LLaVA.}
    \label{tab:4}
\end{table}

 \paragraph{Outer Inconsistency}
 In this context, outer inconsistency refers to models responds `no' to prompt II (because prompt II sets $n$ equaling to its own answer to the direct asking question). If it says this answers is incorrect, it conflicts with itself, which is inconsistent. As can be seen from the `yes' ratio of prompt II, although model generally tends to agree with its own answers (prompt II), there are still notable instances where the model disagree with its own answers. In fact, models have an average inconsistency rate of 16.5\% with its own answers except InstructBLIP. For InstructBLIP, it answers `yes' to almost every question, and interestingly it still has the lowest `yes' ratio of prompt II among the three prompts. This outer inconsistency further reveals the model's uncertainty about the counting problem and may contribute to number hallucination.

\subsection{Comparison Questions}
\label{sec:4.2}
In addition to directly querying the number of a certain object, assessing the model's ability by comparing the numbers of two objects can also provide insights. Comparison is a task that shares both similarities and differences with direct questioning. To some extent, it is more complex as it involves counting two different objects and comparing the results. However, since the comparison task does not necessitate the model to know the exact number of any object, it cannot fully represent the model's counting ability.

Given an image and two objects $c_{1}$ and $c_{2}$, the corresponding comparison question can be formatted in the following two ways. By analyzing the output of both prompts and the primal task, we can also observe inner and outer inconsistency.

\begin{itemize}
    \item I: \textit{Prompt: Which object has a larger number in this picture, $c_{1}$ or $c_{2}$? If they have the same number, answer same.}
    \item II: \textit{Prompt: Please select the correct option according to the given picture. A. There are more $c_{1}$ than $c_{2}$ in this picture. B. There are more $c_{2}$ than $c_{1}$ in this picture. C. The number of $c_{1}$ and $c_{2}$ are the same in this picture.}
\end{itemize}

It can be observed that $c_{1}$ and $c_{2}$ are symmetric in this format. To mitigate the influence of order and analyze inner consistency, each question is repeated with $c_{1}$ and $c_{2}$ flipped. The results is shown in Table \ref{tab:5} \footnote{InstructBLIP cannot response meaningful answers to prompt II, so the corresponding result is not shown.}.


\begin{table}[htbp]
    \centering
    \setlength{\tabcolsep}{5mm}{
    \begin{tabular}{c|cc}
\hline
    Model & Prompt & Accuracy \\
    \hline
    \multirow{2}{*}{LLaVA} & I & 0.456 \\
    ~ & II & 0.368 \\
    \hline
    \multirow{2}{*}{LLaVA-L} & I & 0.478 \\
    ~ & II & 0.412 \\
    \hline
    \multirow{2}{*}{InstructBLIP} & I & 0.377  \\
    ~ & II  & - \\
    \hline
    \multirow{2}{*}{MiniGPT-v2} & I & 0.448 \\
    ~ & II & 0.444 \\
\hline
    \end{tabular}}
    \caption{The results of comparison questions. Accuracy is calculated according to ground truth.}
    \label{tab:5}
\end{table}

\begin{table*}[htbp]
    \centering
    \begin{tabular}[t]{c|c}
    \toprule[1.5pt]
    Input Picture & Prompt and Output \\
    \midrule[1pt]
    \multirow{2}{*}{\includegraphics[width=3.5cm, height=1.9cm]{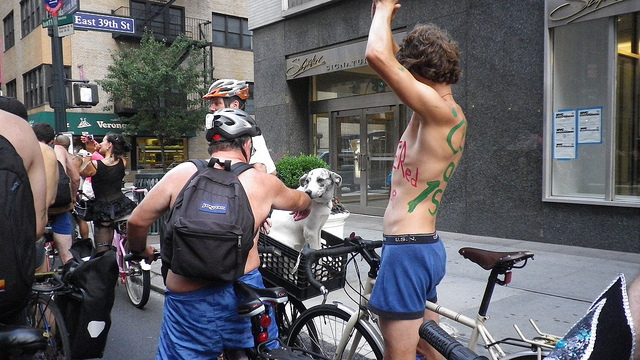}} & 
    \makecell[c]{Prompt: Which object has a larger number in this picture, backpack or dog? \\If they have the same number, answer same. \\ Answer: \textcolor{red}{Dog}} \\ \cline{2-2} ~ & \makecell[c]{Prompt: Which object has a larger number in this picture, dog or backpack? \\If they have the same number, answer same. \\ Answer: \textcolor{green}{Backpack}}\\ 
        \midrule[1.5pt]
        \multirow{2}{*}{\includegraphics[width=3.5cm, height=2.5cm]{latex/fig-app1.jpg}}
         & \makecell[c]{Prompt: Please select the correct option according to the given picture. \\A. There are more dogs than bicycles in this picture.\\ B. There are more bicycles than dogs in this picture.\\ C. The number of dogs and bicycles are the same in this picture. \\ 
         Answer: \textcolor{green}{B}} \\ \cline{2-2}  ~ & \makecell[c]{Prompt: Please select the correct option according to the given picture. \\A. There are more bicycles than dogs in this picture.\\ B. There are more dogs than bicycles in this picture.\\ C. The number of bicycles and dogs are the same in this picture.  \\Answer: \textcolor{red}{B}} \\ 
        \hline
    \end{tabular}
    \caption{Examples of model's response changing after the order of objects (options) changes. The upper case shows one example of inner inconsistency of prompt I, while the lower case shows one example of inner inconsistency of prompt II. Both cases are drawn from MiniGPT-v2.}
    \label{tab:app1}
\end{table*}

As the results demonstrate, all models perform poorly on this comparison task, aligning with our expectations given the earlier findings that all models suffer from severe number hallucination. As a comparison, if the model only guesses an answer randomly, it will still achieve an accuracy of about $0.33$. What's more, a model that always responds with a particular pattern (e.g., always answering `A') will achieve an accuracy of even more than $0.40$, so this result is very disappointing. It is natural that models struggle with comparison problems when they exhibit challenges in accurately counting numbers. However, inner and outer inconsistency in the results can still be observed.

\paragraph{Inner Inconsistency}

Inner inconsistency is similar to what occurs in binary classification questions. In this context, inner inconsistency refers to the model providing different answers to the same question. By comparing within the same prompt format, it is surprising to find that after $c_{1}$ and $c_{2}$ are flipped, the model's response sometimes changes. One example can be found in Table \ref{tab:app1}.

Our quantitative result shows that models generally output inconsistent answers for prompt I at a ratio of 25\%, which is non-negligible. Prompt II also experiences this inconsistency even more severely (an example can be found in Table \ref{tab:app1}). For prompt II, the model responds with inconsistent answers of around 80\%, which is disasterous. What's more, if one only see the correct result(like the answers marked in green in) Table \ref{tab:app1}, he will probably fail to recognize the hidden hallucination inside the model, leading to oversight of the severity of hallucination.

This inconsistency reveals two facts. Firstly, the order of input and the format of prompt have a notable influence on the model's output, making it challenging to evaluate the model's performance with multiple-choice questions. Secondly, this inconsistency also reflects the model's uncertainty about this question, which is a potential cause of number hallucination.

\paragraph{Outer Inconsistency}
Outer inconsistency in this context refers to such a phenomenon that the model's response to comparison task conflicts with its response to primal task. For instance, if the model responds there being $1$ cat and $2$ dogs in primal task but it answers there being more cats than dogs in comparison task, we say the model suffers from outer inconsistency.

Based on the definition above, we conduct a quantitative analysis, and it is not surprising to find that models also suffer from outer inconsistency severely on this comparison task. In fact, models only output 32\% consistent answers averagely, which means under most circumstances the model does not completely agree with its own answer. In fact, a model that randomly responds to these questions will still achieve an outer consistency of about 33\%, which is even higher than current results. Although this comparison task is more complicated and requires more complex reasoning, this low consistency is still absolutely a notable problem and should not be completely attributed to the complexity of this task. 

In summary, through the analysis of results from these two perspectives, we observe severe inner and outer inconsistency problems. This inconsistency problem is not limited to any certain task or specific prompt. Instead, inconsistency appears everywhere without any regular pattern. These issues undermine the reliability of evaluation methods based on indirect questions. And, since this inconsistency represents the model's uncertainty and confusion towards this task, we believe inconsistency is a potential contributing factor to number hallucination.

\begin{table*}[htbp]
    \centering
    \setlength{\tabcolsep}{15pt}
    \begin{tabular}{c|cccc}
    \toprule[1.5pt]
        Model & Finetune Method & Marco-F1 & Weighted-F1 & MAE \\
        
        \midrule[1pt]
        \multirow{5}{*}{LLaVA} & None & 0.186 & 0.288 & 1.84 \\
        ~ & Direct & 0.190 & 0.282 & 1.68 \\
        ~ & Consistency(I) & 0.198 & \textbf{0.302} & 1.70 \\
        ~ & Consistency(II) & 0.195 & 0.287 & 1.79 \\
        ~ & \textbf{Consistency(I+II)} & \textbf{0.212} & 0.301 & \textbf{1.65}\\
        \hline
        \multirow{5}{*}{InstructBLIP} & None & 0.201 & 0.300 & 2.35 \\
        ~ & Direct & 0.235 & 0.324 & 1.63 \\
        ~ & Consistency(I) & 0.235 & 0.346 & 1.99 \\
        ~ & Consistency(II) & 0.216 & 0.322 & 1.62 \\
        ~ & \textbf{Consistency(I+II)} & \textbf{0.255} & \textbf{0.364} & \textbf{1.50} \\
        \hline
        \multirow{5}{*}{MiniGPT-v2} & None & 0.195 & 0.293 & 2.05 \\
        ~ & Direct & 0.219 & 0.318 & \textbf{1.61} \\
        ~ & Consistency(I) & 0.225 & \textbf{0.331} & 1.69 \\
        ~ & Consistency(II) & 0.192 & 0.274 & 1.98 \\
        ~ & \textbf{Consistency(I+II)} & \textbf{0.228} & 0.320 & \textbf{1.61} \\
        \bottomrule[1.5pt]
    \end{tabular}
    \caption{Results of finetuned models.}
    \label{tab:7}
\end{table*}

\section{Mitigating Number Hallucination}

\subsection{Method}
The most straightforward method to mitigate number hallucination is directly finetuning of LVLMs using counting questions. However, as analyzed in Section \ref{sec:4}, since the model suffers from inner and outer inconsistency from different perspectives of this task, and this inconsistency is a potential cause of number hallucination, we would like to introduce a consistency training method to further mitigate number hallucination.

To elaborate further, rather than directly finetuning with the task defined in Section \ref{sec:3.1}, combining the two relevant tasks mentioned in Section \ref{sec:4} may provide the model with a more comprehensive understanding of object numbers, leading to improved performance. Combining different tasks may also enhance the consistency among different perspectives, consequently reducing hallucinations.

Formally, given a training dataset $S = \{(i, c, n_{c})\}$, the direct training method constructs a VQA dataset with questions formatted as follows. This method is called \textit{Direct}.

\textit{Prompt: How many $c$ are there in this picture? Answer in a single number.}

\textit{Answer: $n_{c}$}

The consistency training method constructs a VQA dataset with two question formats:

\begin{itemize}
    \item I: \textit{Prompt: There is(are) $m$ $c$ in this picture, is that correct? How many $c$s are there in this picture?} 

    \textit{Answer: Yes(No). $n_{c}$.}
    \item II. \textit{Prompt: Which object is more in this picture, $c_{1}$ or $c_{2}$? How many $c_{1}$ are there in this picture? How many $c_{2}$ are there in this picture?}

    \textit{Answer: $c_{1}(c_{2})$(The number of $c_{1}$ and $c_{2}$ are the same in this picture.) $n_{c1}$. $n_{c2}$.}
\end{itemize}

We construct our consistency training dataset with three settings: using prompt I only(\textit{Consistency(I)}), using prompt II only (\textit{Consistency(II)}) and using both data generated by prompt I and prompt II (\textit{Consistency(I+II)}). 

By utilizing related tasks, we aim at improving the inner consistency of both tasks. We add the output of primal task to improve outer consistency. To be more detailed, since the model are trained to output correct answers to two different tasks simultaneously, the outer consistency is explicitly improved. We suppose by considering both inner and outer consistency, the model's performance will be further improved. 

\subsection{Experiment Setup}
\paragraph{Dataset Construction}
We utilize the MSCOCO2014 \citep{lin2014mscoco} training set as the foundational dataset and construct $S$ in the same manner as described in Section \ref{sec:3.2}. We refrain from performing k-uniform sampling on the training set to avoid bias and retain as much data as possible. Since we have completely no idea about the data distribution in real-life scenarios, avoiding introducing prior biases and retaining as much data is more reasonable. This provides us with 233k pieces of data. For \textit{Direct} method, all these data are used by directly asking like Section \ref{sec:3.1}. 

For \textit{Consistency(I)}, all these data are kept with prompt I, which provides us with 233k pieces of data similarly. The selection $m$ follows the same setting as setting III mentioned in Section \ref{sec:4.1.III}. That is, 50\% of training data sets $m=n_{c}$, while the rest of training data uniformly selects $m$ around $n$. 

As for \textit{Consistency(II)}, 100k pieces of data are randomly selected with prompt II. This selection aims at limiting the total amount of data. By combining these two datasets, we get a dataset with 333k pieces of data, corresponding to \textit{Consistency(I+II)}. Note that all these consistency data are automatically constructed from the whole original MSCOCO2014 training set.

\paragraph{Model}
We continue the experiments on LLaVA, InstructBLIP and MiniGPT-v2.\footnote{LLaVA and LLaVA-L bear the same architecture, so we only finetune LLaVA for simplicity.} For simplicity and efficiency, we freeze the vision encoder as well as the LLM parts and only finetune the alignment part of each LVLM. For LLaVA, the alignment part is a two-layer MLP. For MiniGPT-v2, the alignment part is a linear layer. Regarding InstructBLIP, though the whole alignment part is a Q-former and a linear layer after it, we perform a preliminary experiment and observe that only finetuning the linear layer after Q-former instead of both Q-former and the linear layer offers a better performance on the \textit{Direct} training method, so we also only finetune the linear layer after Q-former in InstructBLIP. Details of our hyper-parameters can be found in Appendix B.

\subsection{Results and Analysis}
The results are presented in Table \ref{tab:7}. Some conclusions can be drawn from this result. Firstly, directly finetuning the model is useful for mitigating number hallucinations and provides a fine result. 

The \textit{Consistency(I+II)} method not only mitigates number hallucination compared with the un-finetuned models, but also outperforms the \textit{Direct} method by 8\% averagely (macro-F1), clearly demonstrating the effectiveness of our approach. This improvement further supports our assumption in Section \ref{sec:4} that inconsistency is a probable cause of number hallucination. 

It is also worth mentioning that, even though the \textit{Direct} method bears a more similar format with the testing scenario while our \textit{Consistency(I+II)} method has a relatively different format, our \textit{Consistency(I+II)} method still demonstrates better performance, which further reveals the importance of improving consistency.

\paragraph{Ablation Study}
We train with different consistency formats to fully reveal the performance of consistency method. \textit{Consistency(I)} outperforms \textit{Direct} on some metrics, which has already shown the effectiveness of consistency method. However, \textit{Consistency(II)} does not perform as well, which we believe is because comparison only is too difficult and far from our primal task, making it less effective when tested with primal task. 

We observe an interesting phenomenon that \textit{Consistency(I)} often offer better weighted-F1 and worse MAE as well as macro-F1, while \textit{Consistency(I+II)} may bear a little bit worse weighted-F1 with much better MAE than \textit{Consistency(I)}. We believe this phenomenon is because of the distribution of our training set. Since we do not perform double k-uniform sampling on our training set as mentioned, there are much more data with small answers rather than large answers, making the model more confident about generating smaller answers after introducing consistency with binary classification questions. But this actually harms the overall performance because it makes larger mistakes when counting more objects. However, when comparison question is introduced, the model is forced to focus on the whole image instead of trying to bypass by generating smaller answers, which result in a small drop in weighted-F1 while an increase in macro-F1 and MAE. This observation also highlights the importance of using these three metrics instead of one particular metric. As can be seen, using three metrics offers a more comprehensive understanding of the model's performance. 

Further, after combining prompt I and II, \textit{Consistency(I+II)} outperforms \textit{Direct} on almost all metrics and also beats \textit{Consistency(I)} and \textit{Consistency(II)} on most metrics, which highlights the effectiveness of our consistency method and the importance of both perspectives. Although either perspective only does not offer good enough results, combining them leads to an excellent result. This observation also meets our assumption that adding more perspectives may further ease the inconsistency problems and resulting in better overall performance.

\paragraph{Advantages}
Our method offers several advantages. Firstly, it solely focuses on data construction and is independent of specific model architecture. This model-agnostic characteristic makes it extendable to almost any large vision-language model (LVLM). As can be seen from our results, this method improves performance across all tested LVLMs.

Secondly, this method operates without the need to finetune the large language model (LLM) part of the model. Some mitigating methods necessitate finetuning the entire alignment and LLM part of the LVLM, which can be expensive and less efficient. In contrast, our method works with relatively small trainable parts, making it a more cost-effective and efficient solution.

Thirdly, this method may be extended to other hallucination-mitigating tasks. From a methodological perspective, most hallucinations typically involve various perspectives, thus consistency can be analyzed, and similar training methods can be applied.

\paragraph{Case Study}
\begin{table}[htbp]
    \centering
    \begin{tabular}{c|c}
    \toprule[1.5pt]
        \multicolumn{2}{c}{\makecell[c]{{\includegraphics[width=0.3\textwidth]{latex/fig-app1.jpg}}}} \\
        \midrule[1pt]
        \multicolumn{2}{c}{\makecell[c]{Prompt: How many bicycles are there in this picture? \\ Answer in a single number.}} \\
        \hline
        \makecell[c]{Answer of \textit{Direct}:\\
        \textcolor{red}{3}.} & \makecell[c]{Answer of \textit{Consistency(I+II)}: \\ \textcolor{green}{5}.} \\
        \hline
    \end{tabular}
    \caption{A successful example of consistency method. The gold answer is $5$. This example is drawn from InstructBLIP.}
    \label{tab:app-9}
\end{table}

Table \ref{tab:app-9} shows a successful example of our consistency method. This is also a case that we use to illustrate inconsistency problem in \ref{sec:4.2}, which further proves the effectiveness of introducing consistency into training. As can be seen, with \textit{Direct} method the model fails to identify the correct number of bicycles, while after introducing \textit{Consistency(I+II)} method the model responds the correct answer. We believe this improvement is led by our consistency method forces the model to think more carefully and comprehensively, and thus reduces uncertainty and confusion.

\section{Conclusion}
In this paper, we introduce a new form of object hallucination - number hallucination, and present a 20k dataset along with evaluation metrics and results to assess number hallucination in LVLMs. Our findings reveal that all LVLMs suffer severely from number hallucinations. Through our analysis, we demonstrate that binary classification questions are insufficient evaluation methods. We further explore inner and outer inconsistency from two perspectives related to number hallucination and identify severe inner and outer inconsistency despite task or prompt, highlighting inconsistency as a potential cause of number hallucination. Building on this analysis, we propose a consistency training method to mitigate number hallucination and observe an averagely 8\% improvement. This method is extendable to almost any LVLM and applicable to various types of hallucination.

\bibliographystyle{ACM-Reference-Format}
\bibliography{main}

\appendix
\section{Analysis of GPT-4V}
Since our test set is rather large, we randomly select 100 pieces of data from our test set to perform a quantitative evaluation of number hallucination on GPT-4V. The evaluation results are listed in Table \ref{tab:gpt4v}.

\begin{table}[htbp]
    \centering
    \begin{tabular}{c|ccc}
    \hline
        Model & Macro-F1 & Weighted-F1 & MAE \\
        \hline
        Random & 0.086 & 0.103 & 3.56 \\
        LLaVA &  0.186 & 0.288 & 1.84 \\
        LLaVA-L & 0.196 & 0.299 & 1.80 \\
        InstructBLIP & 0.201 & 0.300 & 2.35 \\
        MiniGPT-v2 & 0.193 & 0.293 & 2.06 \\
        \textbf{GPT-4V} & \textbf{0.276} & \textbf{0.447} & \textbf{1.17} \\
        \hline 
    \end{tabular}
    \caption{Number hallucination evaluation results of different LVLMs. GPT-4V is evaluated on only 100 pieces of data.}
    \label{tab:gpt4v}
\end{table}

\begin{figure}
    \centering
    \includegraphics[width=0.3\textwidth]{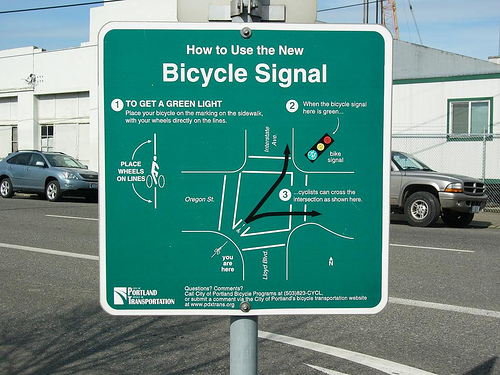}
    \caption{An example of number hallucination in GPT-4V.}
    \label{fig:gpt4v}
\end{figure}

As can be seen from Table \ref{tab:gpt4v}, GPT-4V outperforms other models a lot on all metrics, which meets our expectation that GPT-4V is more powerful and bears less hallucination. However, GPT-4V still suffers from number hallucination, and an example is shown in Figure \ref{fig:gpt4v}.

The prompt is `How many cars are there in this picture? Answer in a single number.' It is obvious that there are only 2 cars in this picture, however GPT-4V responds with 4. This example and the quantitative evaluation results clearly demonstrate that even an LVLM as powerful as GPT-4V still suffers from number hallucination, which further emphasizes the importance of discussing and addressing this problem.

\section{Hyper-Parameters of Finetuning}
\label{sec:hyper}
We finetune each model for three epochs, which costs from $18$ to $30$ hours depending on different models and methods. The remaining hyper-parameters generally remain unchanged as those provided in the official repository of each model. Batch size is modified to fit the model into our computing resources. Gradient accumulation is used to imitate larger batch size.

\begin{table}[htbp]
    \centering
    \begin{tabular}{c|c}
    \hline
        Learning Rate &  1e-5 \\
        Weight Decay & 0.05 \\
        Batch Size & 16\\
        GPUs & 4 \\
        DeepSpeed & zero1 \\ 
        Seed & 42 \\
    \hline
    \end{tabular}
    \caption{Main hyper-parameters used for tuning LLaVA.}
    \label{tab:hyp-llava}
\end{table}

\begin{table}[htbp]
    \centering
    \begin{tabular}{c|c}
    \hline
        Learning Rate & 1e-5 \\
        Weight Decay & 0.05 \\
        Batch Size & 8\\ 
        GPUs & 2 \\
        Seed & 42 \\
    \hline
    \end{tabular}
    \caption{Main hyper-parameters used for tuning InstructBLIP.}
    \label{tab:hyp-inblip}
\end{table}

\begin{table}[htbp]
    \centering
    \begin{tabular}{c|c}
    \hline
        Learning Rate & 1e-5 \\
        Weight Decay & 0.05 \\
        Batch Size & 24 \\
        GPUs & 3 \\
        Seed & 42 \\
    \hline
    \end{tabular}
    \caption{Main hyper-parameters used for tuning MiniGPT-v2}
    \label{tab:hyp-minigpt}
\end{table}



\end{document}